# Swa-bhasha Resource Hub: Romanized Sinhala to Sinhala Transliteration Systems and Data Resources

**Deshan Sumanathilaka[1,2], Sameera Perera[2], Sachithya Dharmasiri[2],**
Maneesha Athukorala[2], Anuja Dilrukshi Herath[2], Rukshan Dias[2] , Pasindu Gamage[2],
Ruvan Weerasinghe [2,3], Y.H.P.P. Priyadarshana [4]
[1]Swansea University, Wales, United Kingdom
[2]Informatics Institute of Technology, Colombo, Sri Lanka
[3] School of Computing, University of Colombo, Colombo, Sri Lanka
[4]Kyoto University of Advanced Science (KUAS), Kyoto, Japan

## Abstract

The Swa-bhasha Resource Hub provides a comprehensive collection of data resources and algorithms developed for Romanized Sinhala to Sinhala transliteration between 2020 and 2026. These resources have played a significant role in advancing research in Sinhala Natural Language Processing (NLP), particularly in training transliteration models and developing applications involving Romanized Sinhala. The current openly accessible datasets and corresponding tools are made publicly available through this hub. This paper provides a detailed overview of the authors' resources and a comparative analysis of existing transliteration applications in the domain.

## 1 Introduction

With the advent of web 2.0, multilingual compatibility in social media and digital platforms has been explored. Sinhala, a low-resourced language, has been used by nearly 13 million Sri Lankans for their communication in both digital and non-digital environments (Perera and Sumanathilaka, 2025b). Romanized Sinhala, commonly referred to as "Singlish", is considered to be an informal way of presenting Sinhala using the Latin script. This writing style has been widely used, largely due to the familiarity with English keyboards and the convenience it offers on digital communication platforms. However, the informal nature of Romanized Sinhala presents significant challenges for natural language processing applications, particularly in the crucial task of backward transliteration, the process of converting Romanized text back into native Sinhala script (Liwera and Ranathunga, 2020). Singlish is non-standardized, leading to ad-hoc typing patterns, spelling variations, and structural inconsistencies, often involving vowel omissions or phonetic adaptations. A central and persistent issue arising from this is lexical ambiguity, where a single Romanized word can correspond to multiple distinct Sinhala words with different meanings depending on the context. For instance,"Adaraya" could transliterate to ආදරය (love) or ආධාරය (aid). Resolving such context-dependent ambiguities is critical, making backward transliteration a greater challenge than forward transliteration, as it necessitates deep contextual awareness. Different romanized patterns and their Sinhala representation are presented in Table 1.

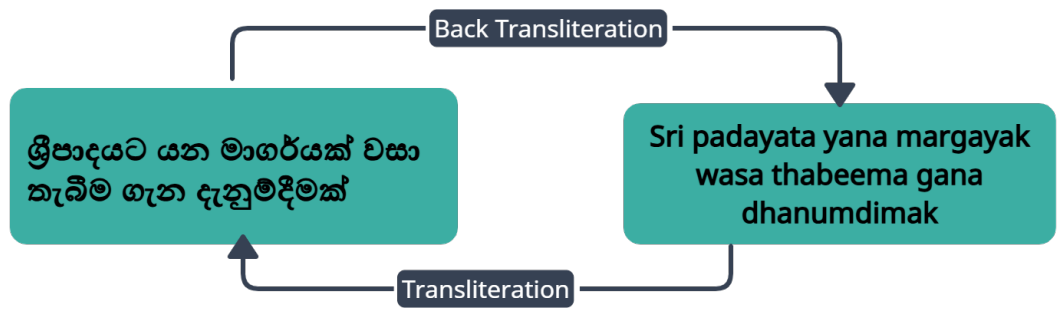

Figure 1: Transliteration vs Back Transliteration

Despite the growing need for robust solutions in this domain, Sinhala, being a low-resource language, faces a scarcity of dedicated linguistic resources and benchmarks specifically designed to evaluate the resolution of transliteration ambiguities through contextual understanding (De Silva, 2019). This collection of papers, representing the work of young NLP enthusiasts, aims to address these critical gaps by exploring and introducing novel context-aware approaches and contributing essential datasets. The work presented herein focuses on developing effective solutions for Romanized Sinhala back-transliteration, with a particular emphasis on handling ad-hoc typing patterns and resolving lexical ambiguities using advanced NLP techniques and purpose-built evaluation resources, thereby advancing NLP for low-resource languages.

The main contribution of this work include:

| Romanized Sinhala | Sinhala | Ad hoc Romanized Sinhala |
|---|---|---|
| Yakoo meeka hindi sindu kiyanawane adenna. Mu indiyawe hitiyanam patta porak welaa aniwaa | යකූ මේක හින්දි සින්දු කියනවනෙ අඩෙන්න. මු ඉන්දියාවෙ හිටියනම් පට්ට පොරක් වෙලා අනිවා | yku mka hindi sindu kiynwne adena. mu indiywe hitiynm ptta pork wela aniw |
| sanka mama dakapu thawath nihathamani soduru minisek | සන්ක මම දැකපු තවත් නිහතමානි සොදුරු මිනිසෙක් | snka mma dkpu twth nihthmni soduru mnisk |
| oba wada satahana karana widiyata bahutharaya kamathiy.ada lankawe niwedakayek wenna ona kenekta puluwan. habay oba wage niwedakayek wenna hamotama ba | ඔබ වැඩ සටහන කරන විදියට බහුතරය කැමතියි.අද ලන්කාවෙ නිවේදකයෙක් වෙන්න ඕන කෙනෙක්ට පුලුවන්. හැබැයි ඔබ වගේ නිවේදකයෙක් වෙන්න හැමෝටම බෑ | oba vde sthna krna widhiyta bahuthrya kmathiy. ada lankwe niwedkyek wenna ona kenekṭa puluwn hbai oba wage niwedkyk wenna hamoṭma ba. |

Table 1: Romanized Sinhala, Sinhala Script, and Ad-hoc Romanized Sinhala

- Summarize the related algorithms associated with Romanized Sinhala to Sinhala by Swabhahsa Team.
- Discuss the implemented Romanized sinhala to sinhala datasets by the team.
- Releasing all the algorithms and data sources as open source to future researchers.

Moving forward, this paper will discuss about the related works by swa-bhasha team, current results on the available transliterations, available datasets and resources by the team and ongoing projects.

## 2 Related Swa-Bhasha Works

### 2.1 Swa Bhasha: Message-Based Singlish to Sinhala Transliteration

*Swa Bhasha: Message-Based Singlish to Sinhala Transliteration* system is primarily rule-based, enhanced by fuzzy logic techniques to facilitate the transliteration of Singlish words into native Sinhala (Athukorala and Sumanathilaka, 2024).

The implementation was carried out using Python within the Jupyter Notebook interface, part of the Anaconda Navigator ecosystem. The front-end Graphical User Interface (GUI) was developed using the Tkinter framework, while natural language processing and machine translation components leveraged the Natural Language Toolkit (NLTK). The fuzzywuzzy library was utilized for string matching, enhancing the system's ability to associate user input with similar words in the Sinhala language dataset. A custom dataset comprising frequently used native Sinhala words was created and numerically encoded to support the transliteration process.

The core functionality of the *Swa Bhasha* system comprises two principal stages: input processing with tokenization and letter mapping, followed by rule mapping and dataset matching. Upon receiving a Singlish input word, the system tokenizes the word into individual characters, which are stored in an initial list. These characters are then matched against a predefined letter dictionary that maps English letters to unique numeric values. The corresponding character-value pairs are stored as tuples in a second list, from which a third list is derived by extracting and preserving the order of numeric values. This third list serves as the input for the subsequent rule mapping phase.

The rule-based engine first detects the presence of vowels by examining the numeric values in the third list. If all values exceed a defined threshold, the word is categorized as containing no vowels (consonant-only). Otherwise, the word is considered either vowel-rich or exhibiting a mixed condition, where vowels may be partially reduced. For consonant-only words, the system calculates the total character count and branches into predefined conditions based on the number of characters, ranging from one to nine. Each condition corresponds to a set of vowel insertion rules derived from Sinhala phonetic patterns, which guide the generation of new candidate lists where hypothetical vowel positions are filled. These candidate values are then concatenated to produce possible transliterations; for example, the Singlish input `khmd` can be expanded to `kohomada`. If vowels are present or the word is in a mixed state, the input bypasses the vowel inference stage and proceeds directly to the similarity matching process. Here, the concatenated candidate values are matched against the native Sinhala word dataset using the `fuzzywuzzy` library. The matching process ranks candidates, and the most probable native Sinhala words are returned. For instance, inputs such as `kiyanna`, `kianna`, or `kiynna` are all accurately mapped to the Sinhala word කියන්න.

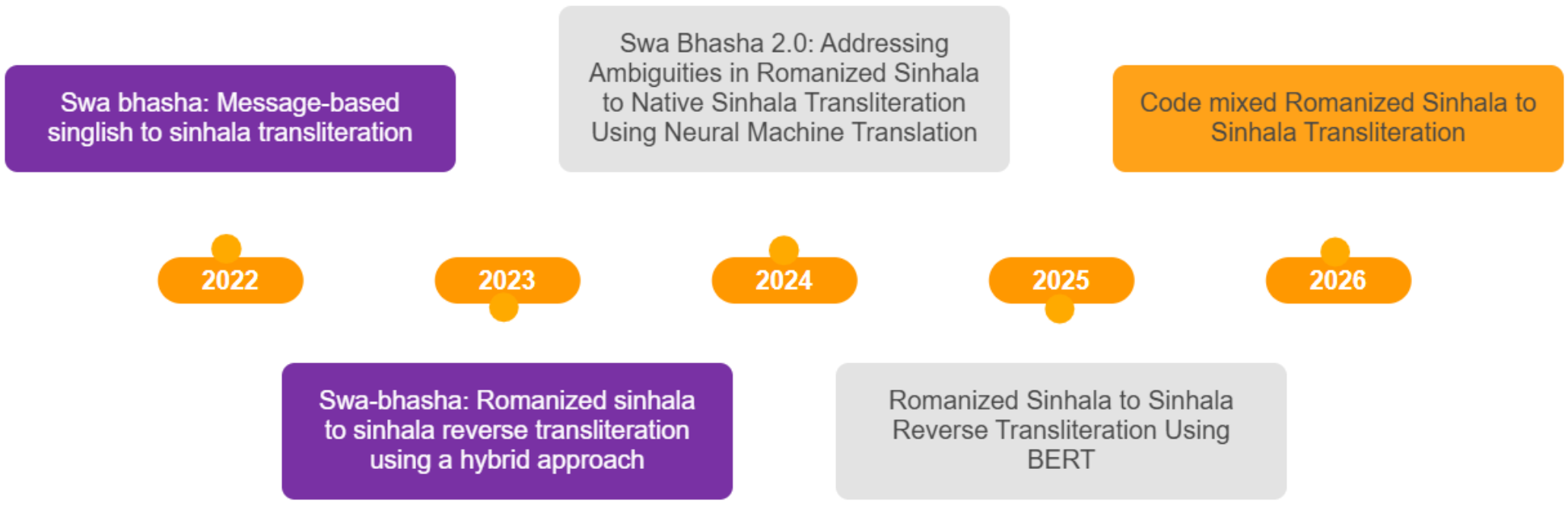

Figure 2: Swa-bhasha Product Development Timeline

The evaluation encompassed three distinct input conditions: words without vowels, words with vowels, and words with reduced vowel counts. Notably, *Swa Bhasha* outperformed comparable transliteration tools such as Google Input Tools and the Helakuru application, particularly in accurately handling vowel-absent Singlish inputs. This underscores the effectiveness of its combined rule-based vowel inference and fuzzy matching methodology.

### 2.2 Swa-Bhasha: Romanized Sinhala to Sinhala Reverse Transliteration using a Hybrid Approach

The methodology proposed in the *Swa-Bhasha: Romanized Sinhala to Sinhala Reverse Transliteration using a Hybrid Approach* (Sumanathilaka et al., 2023) paper integrates a hybrid approach combining statistical, rule-based, and knowledge-based techniques to enhance the transliteration accuracy from Romanized Sinhala (Singlish) to native Sinhala . This is especially relevant for informal communication scenarios where non-standard, shorthand typing patterns are common. The architecture also incorporates a Trie data structure to provide efficient and context-aware word suggestions.

For data collection, the authors compiled a primary dataset comprising Romanized Sinhala sentences and their corresponding Sinhala translations, primarily sourced from social media platforms. To improve transliteration coverage and model robustness, the Dakshina dataset known for its Romanized-Sinhala word pairs was included as a supplementary resource. Additionally, the researchers curated self-composed data from YouTube comments across various domains such as social issues, music, politics, news, religion, and technology. These sentences, containing informal Romanized Sinhala text, were manually transliterated into native Sinhala, thereby forming a high-quality ground truth dataset.

In the data analysis phase, the authors conducted online surveys with 215 participants to study Romanized Sinhala typing patterns. These patterns were extracted using segmentation and alignment methods and then analyzed to derive linguistic rules for data annotation. To examine the evolution of typing behaviors over time, a subset of the original participants was re-engaged. Stratified sampling was employed to ensure that the dataset reflected a diverse demographic, capturing a broad spectrum of real-world typing habits.

Dataset annotation was performed using a custom algorithm developed from the previously analyzed typing patterns. The annotation process generated transliteration pairs by systematically applying rules for vowel placement, consonant-vowel interaction, and disambiguation in cases where a single English consonant could represent multi-ple Sinhala characters. A rule-based transliterator was implemented using 60 transformation rules for vowels and consonants, 18 rules for hal markers, and 18 for special characters. Furthermore, an ad-hoc transliteration generator was developed, which applied an additional 12 character-pattern rules, 6 vowel combination rules, and 8 special character rules. This generator produced an Ad-hoc Romanized Sinhala Dictionary that served as a foundational resource for training the knowledge-based component of the system. To strengthen the dictionary and increase lexical coverage, a Sinhala lexicon developed by the NLP Society of the University of Colombo was integrated into the dataset [1].

For the statistical model, an N-gram tagging system was trained using a comprehensive corpus that included data from the Liwera dataset, the Dakshina dataset, and the self-composed YouTube corpus. In total, the training corpus comprised 12,447 sentences and over 7 million words. The statistical implementation utilized the Natural Language Toolkit (NLTK), employing a backoff-based approach using the UnigramTagger, BigramTagger, and TrigramTagger to maximize tagging accuracy across varying contexts. Annotated output from the rule-based annotation algorithm was also incorporated into the statistical training pipeline.

In the post-processing phase, a secondary rule-based transliteration module was introduced to resolve tokens labeled as NNN, indicating a failure in statistical tagging. These tokens were reprocessed using the ad-hoc transliteration schema and passed through a set of refined rules to generate accurate Sinhala outputs. This dual-layer system ensured that both common and edge-case transliterations were adequately handled.

The final layer of the hybrid approach involved a knowledge-based suggestion system built upon a Trie data structure, implemented as a complete binary tree in Python. This Trie was trained on the Romanized Sinhala patterns generated during annotation and was capable of producing intermediate transliteration suggestions. These suggestions were matched against a knowledge base comprising several domain-specific dictionaries of Romanized-Sinhala mappings. The Trie model returned all potential transliterations, which were then filtered using the knowledge base to produce the most contextually appropriate Sinhala word suggestions. This functionality significantly improved usability by resolving ambiguity and offering intelligent, user-friendly output options. The implementation can be accessed through hugging face [2]. Also the word suggester module is released along with this work [3].

This approach has been further verified and generalised for Tamil. The Tamzi Transliterator has been implemented and tested. Tamzi transliterator uses N-gram analysis, a rule-based model, and a trained trie structure to effectively resolve ambiguities, particularly when vowels are omitted in Romanised Tamil (Herath and Sumanathilaka, 2024).

### 2.3 Swa Bhasha 2.0: Addressing Ambiguities in Romanized Sinhala to Native Sinhala Transliteration using Neural Machine Translation

The paper titled *Swa Bhasha 2.0: Addressing Ambiguities in Romanized Sinhala to Native Sinhala Transliteration* (Dharmasiri and Sumanathilaka, 2024) introduces a novel hybrid methodology designed to accurately and consistently transliterate Romanized Sinhala (commonly known as Singlish) into native Sinhala script. This methodology specifically targets the ambiguities arising from informal and shorthand typing patterns often observed in Singlish. Unlike conventional approaches that rely solely on either rule-based or statistical models, this work combines rule-based techniques with neural machine translation (NMT) enhanced by a suggestion algorithm, aiming to leverage the strengths of both paradigms. The hybrid approach capitalizes on the ability of rule-based systems to encode linguistic knowledge while utilizing NMT models to capture complex contextual relationships and patterns directly from data, thus addressing challenges posed by informal transliterations lacking vowels or using acronyms.

A comprehensive dataset was developed, including a curated set of 44,000 native and Romanized Sinhala word pairs cleaned of stop words, symbols, and duplicates, as well as a larger self-composed corpus of over 50,000 paired terms used for training and validation, divided in an 80:20 ratio. Additionally, an extensive vocabulary consisting of approximately 7 million words was assembled to support the suggestion algorithm.

The rule-based module was implemented using

[1] https://ltrl.ucsc.lk/

[2] https://huggingface.co/deshanksuman/SwaBhasha_Romanized_Sinhala2Sinhala

[3] https://huggingface.co/deshanksuman/Sinhala_Word_Suggestor_for_Singlish

Python and encoded transliteration rules grounded in phonetic and linguistic principles; however, this module alone yielded limited performance, due to the informal nature of Singlish. The advanced NMT component utilized a Seq2Seq architecture with an attention mechanism and employed Gated Recurrent Units (GRUs) to balance training efficiency with the capacity to model long-term dependencies. Trained over 10 epochs using TensorFlow and Keras, the NMT model showed substantial improvement, reducing loss.

To further mitigate word selection ambiguity, a knowledge-based suggestion algorithm employing a suffix tree structure was integrated, drawing from the large vocabulary to present users with contextually relevant alternatives during transliteration. This data-driven hybrid system thus departs from traditional statistical methods, effectively handling the nuances of Singlish transliteration.

### 2.4 Romanized Sinhala to Sinhala Reverse Transliteration Using BERT

The paper titled *Romanized Sinhala to Sinhala Reverse Transliteration Using BERT* (Perera et al., 2025) presents a novel context-aware back-transliteration system designed to address the challenges posed by ad-hoc typing patterns and lexical ambiguity inherent in Romanized Sinhala, commonly known as *Singlish*. The system employs a hybrid methodology that combines multiple components to accurately convert Romanized Sinhala text into native Sinhala script, effectively handling informal variations such as vowel omissions and irregular spellings.

The transliteration process begins with word separation, wherein an input Singlish sentence is segmented into individual words to enable word-level processing. Each Singlish word is then mapped to its corresponding Sinhala equivalents using an ad-hoc transliteration dictionary. This dictionary contains informal Singlish words alongside multiple potential Sinhala translations to account for the inherent ambiguity of Romanized Sinhala. In cases where a Singlish word is not present in the dictionary, a fallback rule-based transliteration approach is applied. This rule-based method uses predefined phonetic mappings, consonant-vowel combinations, and special Sinhala script modifiers to generate a plausible transliteration.

Following this initial mapping, the system assembles a preliminary Sinhala sentence by concatenating the transliterated words. For Singlish words that correspond to multiple Sinhala candidates, the system replaces the ambiguous words with a special `[MASK]` token in the sentence. This token indicates the need for contextual disambiguation, and the set of all possible candidate words for each mask is retained for subsequent processing.

The core of the methodology lies in a context-aware lexical disambiguation phase utilizing a BERT-based masked language modeling approach. For each Sinhala sentence containing masked tokens, the system generates all possible sentence candidates by replacing each mask with every combination of its corresponding Sinhala candidate words. These candidate sentences are then scored using a BERT model fine-tuned on Sinhala text. To compute the score of a sentence, each candidate word is evaluated in context by temporarily masking it in the sentence and passing the partially masked sentence through BERT. The model produces logits for the masked position, which are then converted into a probability distribution using the softmax function. The probability assigned to the correct candidate word reflects how contextually appropriate it is. The overall sentence score is calculated by multiplying the probabilities of each individual candidate word fitting its position. The sentence with the highest score is selected as the final transliteration. For this, the authors employed a pre-trained Sinhala BERT model from Hugging Face, further fine-tuned using Sinhala-language data from the Dakshina dataset, which enhanced transliteration performance.

A key challenge faced in this approach is the computational cost associated with processing long sentences containing multiple ambiguous words, as each additional mask exponentially increases the number of BERT inference calls. To mitigate this issue, two optimization strategies were implemented. First, candidate word reduction was performed by filtering out words not recognized in the BERT tokenizer vocabulary, thereby pruning the search space. Second, for sentences containing three or more masked tokens, a chunking mechanism was introduced. The sentence is divided into overlapping segments, each containing a limited number of masks to keep BERT calls manageable. This chunking ensures that each segment requires fewer BERT calls, reducing processing time to about one second per chunk while maintaining sufficient contextual information for accurate disambiguation.

The system's effectiveness was validated through evaluation on test datasets, where it achieved high BLEU scores alongside low Word Error Rate (WER) and Character Error Rate (CER). These metrics confirm the robustness of the approach in handling both ad-hoc, vowel-omitted Romanized Sinhala and more conventional transliteration cases, demonstrating its suitability for practical applications in Sinhala text input systems. The implementation can be found in the GitHub [4].

### 2.5 Experimental Models for Romanized Sinhala to Sinhala Reverse Transliteration

We developed a series of experimental models based on the facebook/mbart-large-50-many-to-many-mmt multilingual sequence-to-sequence model[5] and a XLMR-roberta from FacebookAI [6]. The initial model was fine-tuned on the Swa-Bhasha transliteration dataset[7], while a subsequent version was trained using two-thirds of the Swa-Bhasha augmented transliteration dataset[8]. These models primarily simulate general Sinhala-to-Romanized writing patterns, as opposed to more ad hoc or idiosyncratic forms of transliteration. Nevertheless, they provide a strong foundation for further fine-tuning on downstream tasks such as shorthand Romanized Sinhala typing.

The trained models are publicly available on the Hugging Face platform: the base model[9] and the model fine-tuned on the augmented dataset[10]. The models were trained using the following hyperparameters: a learning rate of 5e-5, batch size of 32, a maximum sequence length of 128 tokens, and the AdamW optimizer. The number of training epochs was 1. While the current models demonstrate promising results, further improvements are necessary to better accommodate the wide variability in transliteration styles present in informal Sinhala typing.

We developed a custom tokenizer [11] tailored for Romanized Sinhala. This tokenizer is based on the byte-pair encoding (BPE) approach employed by the mBART model and was specifically trained on the Swa-Bhasha Romanized Sinhala Dataset[12]. It introduces a custom language code, "`si_rom`" to represent Romanized Sinhala, thereby enabling more accurate language-specific tokenization. The tokenizer is fully compatible with sequence-to-sequence models, making it suitable for a range of tasks such as transliteration, translation, and text generation in Romanized Sinhala. This tool plays a crucial role in improving model performance by addressing the unique orthographic and morphological characteristics of Romanized Sinhala that are often overlooked by generic tokenizers.

We gratefully acknowledge the support of the Supercomputing Wales project, which is part-funded by the European Regional Development Fund (ERDF) through the Welsh Government, for providing the computational resources used in this research.

### 2.6 Code-mixed Romanized Sinhala to Sinhala translation and transliteration

As a solution to converting the code-mixed romanized Sinhala text to native Sinhala we have developed a hybrid system that consists of a LoRA fine-tuned Neural Machine Translation(NMT) model and a LLM-based post-editing stage.

The Swa-bhasha mBART50 model[10] has been selected as the base model, considering that the model is capable of performing romanized Sinhala transliteration. For fine-tuning the constructed SinMix2Mono dataset, the training split was used. The LoRA fine-tuned model was used in the pipeline, followed by a regex-based post-processing step where it would resolve complex Sinhala letters with zero-width-space issues.

The post-processed text would move in to the LLM-based post-editing stage. As the initial step on the post-editing stage, word alignment would be generated using the provided code-mixed romanized Sinhala text and translated/transliterated Sinhala text from fine-tuned NMT model. The word alignment would be generated using the rule-based code-mixed word-alignment algorithm, which we have developed. Since the code-mixed romanized Sinhala language follows a sequential approach, it was considered to provide the word alignment to the LLM to give a better understanding.

[4] https://github.com/Sameera2001Perera/Singlish-Transliterator

[5] https://huggingface.co/facebook/mbart-large-50-many-to-many-mmt

[6] https://huggingface.co/deshanksuman/swaBhahsa_XLMR_Transliteration

[7] https://huggingface.co/datasets/deshanksuman/SwaBhasha_Transliteration_Sinhala

[8] https://huggingface.co/datasets/deshanksuman/Augmented_SinhalatoRomanizedSinhala_Dataset

[9] https://huggingface.co/deshanksuman/mbart_50_SinhalaTransliteration

[10] https://huggingface.co/deshanksuman/swabhashambart50SinhalaTransliteration

[11] https://huggingface.co/deshanksuman/romanized-sinhala-tokenizer

[12] https://huggingface.co/datasets/deshanksuman/Swabhasha_RomanizedSinhala_Dataset

As the next step of the post-editing stage, the input code-mixed romanized Sinhala text, NMT model output, word-alignment, and instructional prompt would be provided to an LLM asking for final translation. The Gemini-3 Flash model have been selected as the LLM for this proposing architecture, after conducting some benchmarking on different other SOTA LLMs using the SinMix2Mono golden dataset.

The results and component benchmarking would be discussed in the Table 4 and 5.

## 3 Results and Discussion

### 3.1 Evaluation Metrics for Transliteration System

To assess the performance of our transliteration system, we employed a combination of well-established evaluation metrics: Word Error Rate (WER), Character Error Rate (CER), and BLEU score. These metrics were used primarily for evaluating the INDLONLP dataset, which covers a wide range of transliteration pairs.

- Word Error Rate (WER) measures the percentage of words that are incorrectly predicted, providing a high-level indication of output accuracy. Lower WER signifies better overall transliteration quality.
- Character Error Rate (CER) is a finer-grained metric that assesses the percentage of incorrect characters. It is particularly useful in transliteration where even small spelling deviations can significantly affect correctness.
- BLEU Score (Bilingual Evaluation Understudy) evaluates n-gram overlaps between predicted and reference transliterations. Though originally designed for machine translation, BLEU captures fluency and contextual correctness, making it suitable for assessing transliteration quality as well.

For the Transliteration Disambiguation Dataset, we used the F1 Score, which balances precision and recall to assess the system's ability to correctly resolve multiple possible transliterations for a given input. F1 score is ideal here because it accounts for both the relevance of the transliterations selected (precision) and the system's ability to retrieve all correct forms (recall).

These metrics together provide a comprehensive evaluation framework:

- CER and WER capture exactness at different granularities,
- BLEU offers insight into contextual correctness, and
- F1 score measures decision accuracy in disambiguation tasks.

By using this combination, we ensure both the correctness and contextual appropriateness of the transliterations are rigorously evaluated.

### 3.2 The current progress and advancements in the algorithms

The Table 2 presents the performance of several transliteration systems evaluated using Word Error Rate (WER), Character Error Rate (CER), and BLEU scores on two different test sets.

Perera et al. (2025) achieved the best overall performance, with both their Sinhala BERT and Fine-tuned BERT models consistently delivering low WER ( 0.09) and CER ( 0.02), along with high BLEU scores ( 0.91) across both sets. These results highlight the effectiveness of transformer-based models for high-accuracy transliteration. De Mel et al. (2024) compared rule-based and deep learning (DL)-based approaches. The rule-based system performed poorly (WER 0.67–0.68, BLEU 0.01), while the DL-based model showed significant improvement, especially on Set 2 (WER = 0.2413, BLEU = 0.4384), demonstrating the advantage of learning-based methods over rigid rules. Dharmasiri and Sumanathilaka (2024)'s GRU + Rule-based hybrid model showed moderate performance, with WER around 0.33 on Set 1 and 0.48 on Set 2, and a BLEU score ranging from 0.52 to 0.66. Sumanathilaka et al. (2023)'s N-gram + Rule-based model outperformed the purely rule-based system, achieving reasonable WER (0.23–0.25) and high BLEU scores ( 0.75–0.77), suggesting that combining statistical methods with rule-based logic can yield solid results. As a conclusion, Transformer-based models, especially fine-tuned BERT variants, outperform traditional rule-based and hybrid systems by a wide margin in transliteration tasks. This underscores the value of contextual embeddings and deep learning for handling language-specific nuances in transliteration.

| System | Model | Set | WER | CER | BLEU |
|---|---|---|---|---|---|
| Perera et al. (2025) | Sinhala BERT | 1 | 0.0888 | 0.0203 | **0.9113** |
| | | 2 | 0.0917 | 0.0216 | 0.9084 |
| | Fine-tuned BERT | 1 | **0.0867** | **0.0200** | **0.9133** |
| | | 2 | **0.0903** | **0.0215** | **0.9099** |
| De Mel et al. (2024) | Rule-based | 1 | 0.6689 | 0.2119 | 0.0177 |
| | | 2 | 0.6809 | 0.2202 | 0.0163 |
| | DL-based | 1 | 0.1983 | 0.0579 | 0.5268 |
| | | 2 | 0.2413 | 0.0789 | 0.4384 |
| Dharmasiri and Sumanathilaka (2024) | GRU + Rule-based | 1 | 0.3323 | 0.0827 | 0.6677 |
| | | 2 | 0.4808 | 0.1567 | 0.5193 |
| Sumanathilaka et al. (2023) | N-gram + Rule-based | 1 | 0.2342 | 0.0542 | 0.7667 |
| | | 2 | 0.2509 | 0.0678 | 0.7502 |

Table 2: Evaluation results of back-transliterators on the IndoNLP dataset

| System | Model | Set | F1 Score |
|---|---|---|---|
| Perera et al. (2025) | Sinhala BERT | 1 | 0.9601 |
| | | 2 | 0.9209 |
| | Fine-tuned BERT | 1 | **0.9626** |
| | | 2 | **0.9389** |
| Dharmasiri and Sumanathilaka (2024) | GRU + Rule-based | 1 | 0.3422 |
| | | 2 | 0.3312 |
| Sumanathilaka et al. (2023) | N-gram + Rule-based | 1 | 0.3603 |
| | | 2 | 0.3773 |
| Real-Time Unicode Converter (2006) | Rule-based | 1 | 0.3430 |
| | | 2 | 0.3333 |

Table 3: Evaluation results of back-transliterators on the Transliteration disambiguation Dataset

The Table 3 presents F1 scores for various transliteration systems evaluated on two sets of the Transliteration Disambiguation Dataset, which assesses each model's ability to accurately resolve ambiguities in transliterated forms. Perera et al. (2025) demonstrate state-of-the-art performance, with their Sinhala BERT and Fine-tuned BERT models achieving the highest F1 scores across both sets. Sinhala BERT scores 0.9601 (Set 1) and 0.9209 (Set 2), while Fine-tuned BERT improves slightly with 0.9626 (Set 1) and 0.9389 (Set 2). These high scores confirm that contextual embeddings from BERT models are highly effective in disambiguating transliterations. In contrast, hybrid models such as Dharmasiri and Sumanathilaka (2024) 's GRU + Rule-based approach and Sumanathilaka et al. (2023) 's N-gram + Rule-based system show moderate performance, with F1 scores in the 0.33–0.37 range. These models show a slight improvement in Set 2, but lag significantly behind the BERT-based approaches. The Real-Time Unicode Converter (2006), a purely rule-based system, performs similarly to other hybrid rule-based models, with an F1 score of 0.33 on both sets. This highlights the limitations of non-contextual, rule-driven systems in handling transliteration ambiguity. In conclusion, Context-aware deep learning models, particularly transformer-based architectures like BERT, vastly outperform traditional and hybrid rule-based systems in transliteration disambiguation tasks. The results demonstrate the critical importance of contextual understanding for resolving multiple possible interpretations of transliterated input.

### 3.3 Code-mixed Romanized Sinhala translation transliteration

To evaluate the results of the proposed hybrid system that translates transliterate code-mixed romanized Sinhala, the BLEU score, chrF++ score, and BertScore were used as evaluation metrics. From the study conducted for this project, a golden and code-mixed transliteration-ambiguity dataset was constructed for evaluation. The comparative evaluation on the golden dataset has been done across 9 different models, ranging from rule-based and neural transliteration systems to LLM-based translation-transliteration on both zero-shot and few-shot settings, and fine-tuned NMT models to handle code-mixing. The evaluation results have been provided in Table 4.

| Model | BLEU | chrF | BERTScore |
|---|---|---|---|
| *Baseline Systems* | | | |
| Rule-Based | 4.03 | 37.40 | 40.00 |
| SwaBhasha Trigram | 17.35 | 51.65 | 57.00 |
| Sinhala-BERT | 22.28 | 54.31 | 59.00 |
| Sinhala-BERT (finetuned) | 21.64 | 54.05 | 59.00 |
| *Zero-shot LLMs* | | | |
| Qwen-3 Max | 18.47 | 55.87 | 60.03 |
| Gemini 2.5 Flash-Lite | **26.36** | **61.91** | **66.00** |
| *Few-shot LLMs* | | | |
| Qwen-3 Max(few-shot) | 20.57 | **58.11** | 62.00 |
| Gemini 2.5 Flash(few-shot) | **33.43** | 58.07 | **71.00** |
| *Fine-tuned Models** | | | |
| XLM-R | 42.33 | 71.52 | 76.58 |
| M2M100 | 45.66 | 75.44 | 81.11 |
| SwaBhasha-mBART | **49.70** | **78.47** | **81.19** |

Table 4: Evaluation results on the Golden dataset using BLEU, chrF, and BERTScore metrics. *finetuned using SinMix2Mono dataset

While the rule-based and neural transliteration systems comparatively underperformed, the LLM-based and NMT fine-tuned models on SinMix2Mono training dataset performed well. The LLM-based evaluation was performed on both zero-shot and few-shot settings, and it was identified that the few-shot approach worked better, considering the improvement of both the BLEU score and the BertScore. NMT models that were fine-tuned on the SinMix2Mono training dataset have performed the best overall. Compared to those models, the Swabhasha mBART model has performed the best, considering its capability to perform transliteration.

A separate evaluation have been performed against the SinMix2Mono code-mixed transliteration ambiguity dataset. The results have been benchmarked with the evaluation results of same models against the Sinhala transliteration ambiguity dataset compiled by Perera and Sumanathilaka (2025a), which does not contain code-mixed transliteration ambiguity words but only sentences containing Sinhala transliteration ambiguity words. The evaluation and benchmarking results can be found in Table 5.

| Model | BLEU | chrF | BERTScore |
|---|---|---|---|
| *Sinhala Transliteration* | | | |
| SwaBhasha trigram model | 62.88 | 86.91 | 86.00 |
| Gemini 2.5 Flash-Lite | 56.55 | 81.26 | 84.00 |
| XLM-Roberta FT | 76.42 | 92.47 | 91.13 |
| Swa-Bhasha model FT | 77.64 | 92.85 | 92.37 |
| *Code-Mixed Transliteration* | | | |
| SwaBhasha trigram model | 19.92 | 55.03 | 61.00 |
| Gemini 2.5 Flash-Lite | 39.17 | 65.93 | 73.00 |
| XLM-Roberta FT | 44.93 | 70.63 | 74.98 |
| Swa-Bhasha model FT | 49.29 | 74.62 | 78.90 |

Table 5: Evaluation results on the ambiguity translations and transliterations datasets.

From the evaluation results, it can be identified that the proposed Swa-Bhasha LoRA Fine-tuned model outperformed other SOTA models on both Sinhala transliteration ambiguity and code-mixed transliteration ambiguity.

## 4 Datasets and resources

### 4.1 Swa-bhasha Dataset

The **Swa-Bhasha Dataset** was introduced to address the scarcity of resources for Sinhala to Romanized Sinhala transliteration, with a specific focus on supporting diverse *ad-hoc* typing patterns. Sinhala, being a low-resource language, lacks the robust linguistic infrastructure seen in more widely spoken languages, making transliteration particularly challenging. The dataset was designed to fill this gap, offering a comprehensive resource to aid in accurate and real-time transliteration from Romanized Sinhala to native script (Sumanathilaka et al., 2024).

The creation of the Swa-Bhasha dataset followed a multi-stage, rule- and survey-based methodology. The data collection process began by identifying Romanized Sinhala sentences and their corresponding Sinhala script counterparts, primarily extracted from user-generated social media content, especially YouTube comments [13]. This data was further enriched using the publicly available Dakshina dataset from Google, ensuring a diverse transliteration corpus. The compiled dataset spans various thematic domains, including but not limited to social discourse, music, politics, news, religion, and technical content.

To better understand the informal and diverse ways people type Romanized Sinhala, the researchers conducted a two-phase online survey across multiple user communities. The initial survey, conducted over a period of 1.5 months, collected data from 215 participants. It revealed a wide range of informal and inconsistent typing behaviors. A follow-up survey, conducted six months later with a smaller cohort of 25 participants, aimed to assess whether and how individual typing habits evolve over time. The findings highlighted that typing patterns could shift based on temporal factors and user mood. Most participants, primarily aged between 18 and 40, reported using English keyboards on mobile devices to compose Romanized Sinhala messages.

Based on insights from these surveys, a segmentation and alignment analysis was performed to identify variations in writing styles. Using rule-based general transliteration as a foundational model, the researchers developed a set of 92 general rules and 26 special rules to capture common and irregular Romanized Sinhala typing patterns.

These rules formed the basis for a robust data annotation pipeline. A rule-based algorithm was implemented to simulate Romanized Sinhala words across diverse typing styles. This system included 60 rules dedicated to vowel and consonant mappings, 18 rules for "hal" symbol handling, and 18 additional rules to address special characters. These were used to convert Sinhala words into generalized Romanized Sinhala representations.

To generate more nuanced ad-hoc transliterations, a secondary transliteration generator was developed. This module applied 12 character pattern rules, 6 vowel combination rules, and 8 special character rules to simulate realistic, informal user inputs. The output from this process was aligned with a Sinhala dictionary and further validated using the LTRL framework, resulting in a high-quality, linguistically grounded dataset.

Before finalization, the dataset underwent a preprocessing stage to identify and eliminate redundant entries and manually resolve ambiguous word mappings. This quality assurance step ensured the accuracy and reliability of the final resource. The completed Swa-Bhasha corpus comprises **7,134,803** Romanized Sinhala words, generated from **440,024** unique Sinhala root words, effectively capturing the majority of possible Romanized typing variations in the Sinhala language.

The Swa-Bhasha dataset comprises a variety of distinct data types.

- Unique Words / Sinhala - Romanized Sinhala Word-Level Dataset [14]: This component contains a large corpus of Sinhala words and their corresponding Romanized Sinhala representations, designed to cover most possible typing sequences of general Sinhala words. For instance, a single Sinhala word might be mapped to multiple Romanized forms to account for variations like vowel omissions. The corpus specifically includes 7,134,803 (7.13 Million) words generated for 440,024 unique Sinhala words.

- Romanized Sinhala - Sinhala Social Media Sentence Pair Dataset [15]: This part of the dataset includes pairs of Romanized Sinhala sentences and their corresponding Sinhala sentences, primarily sourced from social media content, specifically YouTube. This inclusion helps capture real-world informal communication patterns. The data collection process involved identifying Romanized Sinhala sentences and their matching Sinhala sentences from social media, augmented by the Dakshina dataset.

- Romanized Sinhala Transliteration Dataset for Transliteration Disambiguation (WSD) [16]: This component is specifically designed to aid in resolving ambiguities during transliteration, where a single Romanized form might correspond to multiple Sinhala words depending on the context. The dataset cover most possible sinhala words, for a ad-hoc format of a romanized sinhala word.

[13] `https://www.kaggle.com/datasets/tgdeshank/romanized-sinhala-sinhala-social-media-dataset/`

[14] `https://www.kaggle.com/datasets/tgdeshank/swa-bhasha-dataset`

[15] `https://www.kaggle.com/datasets/tgdeshank/romanized-sinhala-sinhala-social-media-dataset`

[16] `https://www.kaggle.com/datasets/tgdeshank/wsd-romanized-sinhala-dataset`

### 4.2 IndoNLP shared task Evaluation Dataset

For the Shared Task on Real-Time Reverse Transliteration for Romanized Indo-Aryan languages, a standard evaluation dataset was introduced for five languages, including Sinhala (Sumanathilaka et al., 2025) in the first workshop of Natural Language Processing for Indo-Aryan and Dravidian Languages (Weerasinghe et al., 2025). The test dataset specifically for this IndoNLP Shared Task was created and augmented by combining existing publicly available datasets with newly generated data samples to ensure diversity and relevance. For Sinhala, the evaluation involved two distinct test sets [17] :

- **Test Set 1: General Typing Patterns** – Consisted of 10,000 data points.
- **Test Set 2: Ad-hoc Typing Patterns** – Consisted of 5,000 data points.

The design of these test sets, especially the inclusion of *ad-hoc typing patterns*, aims to address key challenges in Romanized Indo-Aryan language transliteration. These challenges arise due to users frequently employing inconsistent spellings, such as omission or substitution of vowels, and phonetic ambiguities, which complicate accurate mapping to native scripts. Social media communication, in particular, often involves abbreviated and informal typing styles. The objective of the shared task was to develop a real-time reverse transliterator capable of handling such ad-hoc transliterations. This includes inputs with inconsistent or missing vowels, abbreviations, and phonetic variations, in order to produce accurate native script output.

### 4.3 Augmented Romanised Sinhala dataset

To further support research in Sinhala language processing, an **Augmented Sinhala–Romanized Sinhala Dataset** was developed [18], comprising approximately **7.24 million sentence pairs** of Sinhala text and their corresponding romanized versions. This dataset was constructed using publicly available Sinhala language resources, including the SemiSOLD corpus (Ranasinghe et al., 2022), the Sinhala Sentiment Analysis Dataset, and the NSINA corpus (Hettiarachchi et al., 2024). The primary objective of this resource is to facilitate tasks that require standardized romanized representations of Sinhala text, such as transliteration, text normalisation, and language modelling.

The romanized versions of the Sinhala sentences were generated using a standardized transliteration process that adheres to established linguistic conventions for mapping Sinhala script to Roman characters. This ensures consistency and reliability in the transliteration output, making the dataset suitable for formal language processing tasks. However, it is important to note that this dataset focuses on standardized transliteration and does not account for the informal or *ad-hoc* typing patterns commonly observed in user-generated content on social media or messaging platforms. As such, while the dataset offers broad coverage and scale, it is intended for use in scenarios where uniform transliteration patterns are required, rather than applications that aim to model informal, user-specific input styles.

### 4.4 A Dataset for Transliteration Disambiguation

The Transliteration Disambiguation (TD) dataset for Romanized Sinhala is the first dataset specifically developed to tackle transliteration ambiguity in backward transliteration tasks. Its primary aim is to evaluate how effectively systems can resolve ambiguities in real-world, context-dependent usage. Romanized Sinhala often lacks a one-to-one mapping with its native Sinhala script, and the informal, inconsistent spelling conventions used by speakers create scenarios where a single Romanized word may correspond to multiple Sinhala words with distinct meanings. This dataset focuses on binary transliteration ambiguity, where each Romanized word maps to exactly two semantically distinct Sinhala words. The dataset was created through a rigorous multi-step process. First, ambiguous words were selected using an existing Romanized Sinhala–Sinhala dictionary. Sinhala spellings were verified using the Madura Dictionary API, and semantically redundant entries were filtered out. Only Romanized words whose two meanings ranked among the top 3,000 most frequent Sinhala words were retained, resulting in a final list of 22 ambiguous words. Next, 30 unique Sinhala sentences were created for each ambiguous word: 10 for each meaning and 10 containing both meanings within a single context. Sentences were sourced from publicly available mate-rials, with additional dual-sense sentences generated using the Claude 3.0 Sonnet language model. These sentences were then transliterated into Romanized Sinhala using a rule-based system, with manual replacement of target ambiguous terms and phonetic consistency throughout. All entries were manually verified by three linguists to ensure accuracy. The final TD dataset comprises 660 sentence pairs, each consisting of a Romanized Sinhala sentence and its Sinhala script equivalent. It is divided into two test sets: one containing single-sense sentences to test systems' ability to infer meaning from context, and another with dual-sense sentences to assess disambiguation performance in semantically dense scenarios. The dataset offers diverse contextual challenges that require systems to go beyond surface forms and utilize sentence-level semantics for accurate backward transliteration.

[17] https://github.com/IndoNLP-Workshop/IndoNLP-2025-Shared-Task/

[18] https://huggingface.co/datasets/deshanksuman/Augmented_SinhalatoRomanizedSinhala_Dataset

The dataset is organized into two distinct test sets:

- Test Set 1 – Single-Sense Sentences: This set contains Romanized Sinhala sentences where the ambiguous word is used in a context corresponding to only one of its possible Sinhala meanings. It evaluates a system's ability to infer the correct meaning based on sentence-level context.

- Test Set 2 – Dual-Sense Sentences: This set contains sentences where both Sinhala meanings of the ambiguous word appear within the same sentence. It assesses the system's capacity to disambiguate and correctly transliterate each instance of the word in a more complex, semantically dense context.

## 4.5 Code-Mixed Romanized Translation Transliteration Dataset

Code-Mixed romanized Sinhala being an under-explored research area, a collection of datasets has been compiled to support downstream Sinhala linguistic research in the domain (Dias and Sumanathilaka, 2025). The dataset, named SinMix2Mono, contains 3 main data splits: a training split, a golden split for general evaluation, and a Code-Mixed transliteration ambiguity split for ambiguity evaluation. The dataset comprises both real-world scraped data and synthetic data spanning a wider domain.

### 4.5.1 Training and Golden Split

Both the training and golden dataset splits have been constructed via web scraping from real-world data on YouTube comments. When collecting the data, to make the dataset more diverse the collection was done on different domains. The data can be divided into 5 main domains; Entertainment, Business and Education, Science and Technology, News and Law/Politics, and General. From the web scraping, around 20,000 data records were collected initially across all of the mentioned 5 domains. In addition to the web-scrapped data more data were collected by filtering the Swa-Bhasha dataset (Sumanathilaka et al., 2024). It was identified that the Swa-Bhasha dataset also contains code-mixed romanized data records that could be used for this task. As the initial annotation process, the collected data were transliterated using the hybrid transliteration model, which combines trigram-based and rule-based models (Sumanathilaka et al., 2023). This model would produce the Sinhala transliteration, which is also capable of handling ad-hoc transliterations. Since the hybrid transliteration model were not able to handle code-mixing, a second annotation stage was conducted using Claude-Sonnet-3.7 and Claude-Sonnet-4 LLMs to perform the translation while preserving the meaning of the sentence. A detailed few-shot instruction prompt was provided to the LLM to make sure the overall output contains quality translation/transliterations. While the LLM produced quality translations and transliterations, there were errors in complex code-mixed lexical patterns, name entity handling, invalid synonyms, etc. Hence, it was necessary to perform manual human-in-the-loop post-editing for each data record.

For the final dataset (25,00 rows), data splitting was performed with 90% for the training set and 10% for the Golden dataset. The golden dataset underwent further annotation and evaluation, with each data record labelled as correct, incorrect, or invalid. This annotation task was carried out with an Inter-Annotator Agreement among 3 Native Sinhala speaking annotators. After the annotation and evaluation process was complete, the feedback and inputs from annotators were closely reviewed. Sentences that were labelled as incorrect were fixed with correct translation/transliteration, and sentences that were labelled as invalid were removed. After the validation process, the annotator agreement was measured using Gwet's AC1, yielding a score of *0.7465* and a raw agreement rate of *71.78%*. Finally, completing the golden dataset of 2,549 data rows.

#### 4.5.2 Code-mixed transliteration ambiguity Split

Compared with the training and golden datasets, the Code-mixed transliteration ambiguity dataset was compiled from synthetically generated data. Initially, 34 words have been identified that can be considered code-mixed transliteration-ambiguity words. Using those words, the sentences have been generated synthetically using Gemini-3 Flash LLM. When synthetically generating the sentences, it has been instructed to provide 15 different sentences per given ambiguous word. The total of 15 would be 5 sentences in the Sinhala context, 5 sentences in the English context, and 5 sentences in both Sinhala and English contexts combined.

From the generation it got received 510 data records in total. As similar in the Golden dataset, in the ambiguity dataset 30% of the dataset has been send to perform validation/annotation under an Inter-annotator Agreement (IAA). The same labels (correct, incorrect, and Invalid) was used with the same intent. The Gwet's AC1 score was also calculated for this task, while receiving *0.7068*. Manual post-editing was conducted considering the feedback from the annotators. Also, proofreading and updating the remaining 70% of the dataset. Finally, both splits have been combined to construct the Code-mixed transliteration ambiguity dataset.

## 5 Ongoing Projects by Swa-Bhasha Team

### 5.1 Romanized Sinhala Next Word prediction Module

There has been limited work specifically focused on predictive text input for Singlish keyboards. While many computational projects have explored transliteration from Singlish to Sinhala, direct word prediction models for Singlish typing remain underdeveloped. The current ongoing project aims to address this gap by developing a personalized, privacy-preserving word prediction framework for Singlish keyboards. Singlish poses several unique challenges for NLP due to its orthographic inconsistency, frequent code-switching, and strong context dependency. Traditional language models struggle with Data sparsity and Personalized usage patterns: Users often develop idiosyncratic blends of Sinhala and English syntax, making generalization difficult.

To tackle these challenges, this project investigates the use of meta-learning techniques, particularly Model-Agnostic Meta-Learning (MAML), which enables models to adapt quickly to new patterns using only a few examples. This approach shows promise in learning user-specific linguistic styles and handling low-resource conditions effectively. The ongoing work involves designing and implementing a meta-learning-based architecture tailored for Singlish word prediction. The goal is to support real-time, adaptive predictions that respect user privacy while addressing the dynamic and evolving nature of Singlish communication.

## 6 Conclusion

The Swa-Bhasha Resource Hub represents a significant advancement in the domain of Sinhala NLP, particularly in the area of Romanized Sinhala to native Sinhala transliteration. By combining rule-based methods, statistical models, and modern deep learning techniques especially transformer-based models like BERT—the research team has achieved state-of-the-art performance in transliteration tasks, as evidenced by low error rates and high BLEU and F1 scores across benchmark datasets.

Despite these promising results, particularly with transformer-based models that excel in handling contextual ambiguities, a substantial gap remains in developing efficient real-time transliteration systems. The computational demands of models like BERT, especially in handling longer and ambiguity-rich sentences, limit their direct applicability in mobile or low-resource environments where real-time responsiveness is critical. Moreover, while extensive datasets have been compiled, including annotated corpora covering informal typing patterns and social media discourse, challenges persist in standardizing transliteration outputs and adapting systems to dynamic code-mixed and ad-hoc user inputs. The team's ongoing efforts to address code-mixed inputs, especially in the Sinhala-English context, are a step in the right direction, but further research is needed to enhance scalability, generalization across domains, and user adaptability.

In summary, the Swa-Bhasha initiative has laid a robust foundation for Sinhala transliteration re-search and tool development. Future work should focus on optimizing inference efficiency, integrating transliteration systems with broader NLP pipelines (e.g., sentiment analysis, chat interfaces), and ensuring equitable accessibility to these tools in real-world, multilingual digital ecosystems.

## Acknowledgement

We would like to express our sincere gratitude to the reviewers of all related papers for their valuable feedback and constructive suggestions, which greatly contributed to improving the quality of this work. Our heartfelt thanks go to the Helakuru team for providing a detailed repository on Sinhala transliterations. Their resources were instrumental in the data creation process, particularly in typing Sinhala Unicode characters efficiently. We also extend our appreciation to all annotators and Sinhala dataset creators, whose efforts enabled us to build an augmented dataset essential for this study. Their contributions have been vital in supporting research in low-resource language processing. Finally, we would like to acknowledge Dr Ruvan Weerasinghe from the University of Colombo School of Computing, Sri Lanka, and Dr Prasan Yapa from Kyoto University of Advanced Science, Japan, for their guidance and support throughout this project. Finally, we would like to acknowledge Ms Harini Guruge for her dedicated effort in identifying ad-hoc typing styles, a contribution that helped shape this series of papers and advance research into a new and important area.